# Bangla Sign Language Translation: Dataset Creation Challenges, Benchmarking and Prospects

Husne Ara Rubaiyeat, Hasan Mahmud, Md Kamrul Hasan
Systems and Software Lab (SSL), Islamic University of Technology (IUT)

**Abstract:**

Bangla Sign Language Translation (BdSLT) has been severely constrained so far as the language itself is very low resource. Standard sentence level dataset creation for BdSLT is of immense importance for developing AI based assistive tools for deaf and hard of hearing people of Bangla speaking community. In this paper, we present a dataset, "*IsharaKhobor*" (ইশারা_খবর), and two subset of it for enabling research. We also present the challenges towards developing the dataset and present some way forward by benchmarking with landmark based raw and RQE embedding. We do some ablation on vocabulary restriction and canonicalization of the same within the dataset, which resulted in two more datasets, "*IsharaKhobor_small*" and "*IsharaKhobor_canonical_small*". The dataset is publicly available at: www.kaggle.com/datasets/hasanssl/isharakhobor [1].

1. Introduction

Sign Language Translation (SLT) has been proven very hard for several reasons. Firstly there are numerous challenges to create a standard dataset in any language, related to sign sentence construction, gloss annotation and the amount of natural language vocabulary compared to sign glosses. The second issue is because of the complex alignment of sign frames to glosses and gloss to sentence. While there are other challenges, researchers are pushing way forward towards end-to-end sign language translation, skipping gloss annotations. Well known SLT datasets [2,3] have been carefully crafted by researchers either by restricting to a particular topic like weather or by choosing daily usage scenario. Still the datasets suffered low translation accuracy on contemporary deep learning Transformer models marked by Bleu-4 score ranging between 20-25. The datasets created from the wild [4,5] like news suffered severely less accuracy because of the number of vocabulary compared to number of samples, and lack of gloss annotation.

Bangla Sign Language (BdSL) has been extremely low resource so far. It is said that "Bangla Sign Language is kept alive by the state-run television BTV". However, it is the only accessible BdSL resource for all, as Bangladesh Television (BTV) broadcasts a news session accompanied by sign translation almost every day, namely "Desh O Jonopoder Khobor" (দেশ ও জনপদের খবর) and has made the videos publicly accessible through youtube. The broadcast covers almost all the topics possible to present daily events of a country, including government, sports, public awareness, forecasting, national disasters, etc. This presents a unique opportunity: anybody can create a BdSLT dataset and the challenges: can the dataset be automatically collected without expert and how to tackle, restrict the immense vocabular. So, the dataset creation process becomes a worthy experience to describe.

In this paper we describe the dataset creation process of "*IsharaKhobor*" (ইশারা_খবর) and benchmarking. We also challenges to create a standard datasets for BdSLT and the prospects.

## 2. Related Works

### 2.1 Datasets

Sign language dataset creation is a delicate and difficult task, as we see there are not many SLT datasets worldwide. Though early works were to engage sign experts to produce small scale dataset [6], the first large scale dataset was derived from German Weather broadcast. PHOENIX-Weather-2014T[2] was the enabling one for sign language researchers to apply deep learning techniques. CSL-Daily[3] and How2Sign[4] followed the path but instead of choosing broadcast, they engaged volunteers guided by reference sign video from sign experts. The accuracy of sign language datasets depend on the language vocabulary assuming it is gloss-free. Table 1 shows the well-known SLT datasets and their accuracy.

Table 1: Well known large scale Datasets and their accuracies

| Dataset | # of Samples | Topic | Language Vocabulary | Source | Baseline BLEU-4 (gloss free) |
|---|---|---|---|---|---|
| PHOENIX-Weather-2014T [2] | 8,257 | Weather | 2,887 (3,004 with OOV) | TV | 20.17 |
| CSL-Daily [3] | 20,654 | Daily Life | 2,343 ( 2,476 with OOV) | Volunteers & Signers | 13.1 |
| How2Sign [4] | 35,191 | How2 | 15,686(16,609 (with OOV) | Volunteers & Signers | 1.25 |
| BBC-Oxford BSL [5] | >2M, (25,045 human verified test samples) | News events | 24.7K | TV | 1.00 |

OOV= Out of Vocabulary

BBC-Oxford BSL [5] described how to collect sign sentences from news archives in details.

BdSLT dataset construction effort was initiated by BTVSL[7], however they were wrongly assuming the sign sentences can be annotated by voice of the news presenter. They pointed out that a big news segment can be cropped based on voice silence for a while. This assumption is also not fully confirmed as signers mostly start late and end late, but early finishing is also possible. Finger spelling at the end of a segment may take arbitrarily more time for the signer to finish.  However, automation is still possible tracing the signer hands-down within offsets before or after the presenters' long pause. But it is important to understand that long segments with 5-10 sentences together are not any good sign samples. Moreover, sentences from the speaker may not fully correspond to the sign video as we pointed in a later section (Dataset Construction).  We resorted to manual verification of transcripts, and relied on sign expert for sign sentence annotations. In a dataset construction capturing the ground truth is all that important.

The sign dataset from Bagla news naturally is gloss-free with high language vocabulary. To achieve higher accuracy, we need gloss annotation of those sentences. Saha et.al. [8] publishes  1000 sentence-gloss pair ; 500 pair were annotated by expert and other 500 were generated by bangla-bert [9] using masked token prediction. There were mostly tense related. The dataset was synthetically extended to 3000, using RAG and prompting technique. Extension of these researches can help achieve gloss annotations for sentence cropped from Bangla news.

If gloss annotations are not available, alternative approaches can be vocabulary standardization or canonicalization. This may reduce language styles but may produce sign translations with higher accuracy. We took an initiative here in this work and made  the datasets and benchmarks publicly available.

**2.2 Models and Embeddings**

Sincan et.al. [10] present a good overview of gloss-free transformers and compare them in the same settings.  As the focus of this work is not in model, we describe the models from conceptual level.

**2.2.1 Sign language Translation Transformer (SLTT)** [11]

SLTT is a sub-model of the joint transformer SLRT & SLTT, where gloss recognition is done by CTC. SLTT runs without gloss supervision. It is fully regression based and supposed to do well provided many samples, which is unlikely in sign language datasets.

**2.2.2 Gloss Attention Sign Language Transformer (GASLT)** [12]

GASLT predicts the gloss width in video frames and uses natural language sentence similarity as gloss similarity to direct the video encoder heuristically. This allows GASLT to converse quickly.

2.2.3 Gloss-Free SLT based on Visual-Language Pretraining (GFSLT-VLP) [13]

GFSLT-VLP provides a clip inspired latent embedding for the sign video and corresponding text sentence. The model proposes pretraining of the visual language semantics so the latent embedding is better learnt.

**2.2.4 Cross-lingual Contrastive learning (CiCo) [14]**

CiCo also inspired by CLIP and similar to GFSLT but differs in video encoding. They form features from 16 frame clip and stack all such features derived by sliding window. The stacked features and sentence encoding are put a contrastive learning like GFSLT-VLP did. The results reported by CiCo are not comparable as it provided R@K results.

**2.2.5 TwoStream-SLR [15]**

**TwoStream-SLR** takes video and its frame-wise key points as heatmap and runs two different encoders for them. The loss functions are guided by gloss CTC loss for each of the encoders and joint CTC loss, frame to gloss alignment loss termed as distillation and Translation loss. The whole work is overly emphasizing gloss alignment. The final loss function is measured as gloss alignment loss and translation loss, which is a particular configuration of joint SLRT and SLTT work [11]. The overly focus on gloss alignment makes the work incomparable in our dataset as IsharaKhobor does not have any gloss annotation.

**2.2.6 LLM and multimodality**

As gloss annotations need experts, gloss free SLT approaches are adapting multimodality. A very recent work [16] has used LLM to generate descriptions on sign positions & motion generated by prompting. Video features and stacked frame level text descriptions and then put to multimodal language pretraining to learn latent embedding with spoken sentence. While is approach is like GFSLT-VLP, it adds multimodality of LLM generated sign descriptions. This is research direction to follow.

However, it is to observe that even with language descriptions and CLIP like latent embedding pretraining they achieved Bleu-4 of 25.73 in PHOENIX-Weather-2014T dataset, where the baseline gloss-free Bleu-4 is 20.17. In CSL-Daily they achieved 21.11, where gloss-free baseline is 13.1.

In a recent work [17] the authors tried to generate sign descriptions and motions for the whole sign video, that are classifiable in a word level dataset. The descriptions though showed promises, but was not quite decerning for classification.

We felt that LLM has an inherent limitation of generating variable responses for the same input. Moreover, LLMs may not be used in SLT with whatever special embedding we got, as they may not keep the spoken sentence as it is that is necessary for aligning the gloss with text. For example if বেতাল (Discordance) is broken as বে and তাল (palm) where the two main words have very different glosses, we can realize what the end translation will be. SLT in LLM can only be successful when we have millions and billions of samples. The same problems are more applicable on a sign task specific transformer.

The missing link is there is no alignment of sign video and any other modality generated by LLM with the gloss, and of course there is a huge gap in alignment gloss free SLT. As we worked with BdSL word/gloss level dataset [18,19], we experienced any fine variation of gesture generates a new gloss that is way semantically different from the gloss we intend.

In this work we benchmarked Isharakhobor with SLTT and GASLT, using mediapipe based landmark and a derived one RQE, developed by us.

### 3. Dataset Construction

The dataset construction started with scrapping videos of "Desh O Jonopoder Khobor" (দেশ ও জনপদের খবর) from youtube. More than 700 videos were downloaded. Then came the need of transcript generation.

**3.1 Transcript Generation & Verification**

An open source Automatic Speech Recognition (ASR) tool that won a competition on Bangla language was fine tuned on the same dataset by removing noise and used to generate the transcript. The transcripts were good but still erroneous. In sign language even character level error is not expected as that change the sign translation completely, for example, when কলা (Banana) becomes গলা (throat). The transcript posed with another interesting problem, regional dialect. In desk report segments some interviewees talked in regional dialect, however, sign professionals glossed them in standard sign language. We realized the need for transcription verification and correction. We employed six transcript verifiers who corrected the transcripts by listening to videos and if necessary correct the regional dialect. If sentences were too noisy to correct, the verifiers discarded those.

We did not opt for commercial ASR for three reasons, the time we generated the transcripts, commercial tools were not matured enough, second regional dialects are not corrected by those tools and time annotations of sentences received for sentences cannot be used in sign video annotations.

The transcript verifiers were fluent typist in Bagla and were appropriately paid. They delivered transcripts for 650 videos. The transcripts were already broken into sentences.

**3.2 Annotation by Sign Expert**

We realized that sign sentences cannot be automatically segmented by voice annotation. Signers mostly start their work after a while when they listen the news presenter. Often the professional signer continues signing after the news segment is over. Occasionally, the signer starts signing before the presenter, when signer knows what sentence is coming next (towards the end of the news). Sentence level alignment is not mostly possible as signer mostly start late. Sometimes signer skips sign of a sentence, specially muslim mourning or the name of the reporter in desk report segment. In such cases the alignment sequence is altered. Sometimes some segment of sign is repeated in next segment, especially for desk report of a preceding news segment. So, obviously text-sign pair is not aligned by voice. So, our enthusiasm had to be proven by hiring a sign expert. The task was not easy. The researchers will understand that if we make a statement that there is only 8 sign professionals to present news in BTV in over 5 years since the broadcastings were made public.

The sign expert was provided with downloaded videos and the sentence level transcripts. The expert was assisted by a computer operator to help mark the start and end of the sign sentence in the video. The sign expert minutely annotated the sign sentences in frame level and returned us sign annotated transcripts for 500 videos. The sign expert and computer operations were professionally paid.

### 3.3 Verification of Transcripts Again

It was noticed that the corrected transcripts had several unintentional and hard to detect mistake like আয়োজন (to organized something) was misspelled as আয়াজন. So the authors corrected those in json files.

### 3.4 Selection of Neutral Sentences

The authors then endeavored to select neutral sentences within the sign annotated files, by removing sentences with political names and detaching them from contexts.

### 3.5 Dataset Cropping

The annotated transcripts were then converted to json files for each video which also included the urls, sign names, frame rate, resolution, signer cropping regions, etc. This allowed us to automatically crop clips for each sign sentence. The clips were named in combination with video name, start and end frame marker. This allowed us to trace video clips back to json attributes.

We release 180 video annotations in this dataset. Cropping resulted in 5642 video clips of cumulative 11.309 hours from 180 news video of 54.48 hours. We provide annotation and clips for one more video as extra. This extra annotations are for testing embedding that work in low resolution.

### 3.6 Dataset Description

We are releasing 180 json files with sentence level sign annotations containing neutral sentence only. We name it "IsharaKhobor" (ইশারা_খবর) which as 5642 annotated sentences with 11359 vocabulary. One more json file of a very low resolution video has been added as challenge for researchers who claim to have robust video embedding. We discarded the low resolution file, as mediapipe landmark extraction failed in this case of clip width <50 pixels.

#### 3.6.1 Dataset Split

We randomly marked split as train-test-val in 7:2:1 in each json file.

#### 3.6.2 Dataset Statistics

Table 2 and Table 3 describe the attributes and some vocabulary statistics.

**Table 2: Dataset Statistics**

| Attribute | Details |
|---|---|
| Sourece | Bangladesh Television (BTV) news 'Desh O Jonopoder Khobor' (দেশ ও জনপদের খবর) |
| Extracted News | 180 (+1 low resolution) |
| Period | Over 4.5 years ( 8 DEC 2017 -19 AUG 2023) |
| Signers | 8 |
| Extracted Video Duration | 11.309 hours |
| Total News Duration | 54.48 hours |
| Train-Test-Val set construction strategy | Random 7:2:1 ratio from each json file |
| Train-Test-Val set samples and ratio | 3964-1108-570 (ratio: 0.70-0.196-0.100) |

**Table 3: Some vocabulary statistics**

| | |
|---|---|
| Highest Occurances | 'ও': 1804, 'করা': 796, 'হয়': 721, 'হয়েছে': 683, 'জেলা': 661, 'গতকাল': 563, 'এ': 549, 'আজ': 498, 'এবং': 462, 'অনুষ্ঠিত': 454, 'থেকে': 442, 'এই': 409, 'এক': 370, 'করে': 339, 'তিনি': 333, 'উপজেলার': 325, 'করেন': 314, 'বিভিন্ন': 292, 'দেশ': 284, 'হাজার': 284, 'বলেন': 283, 'জনপদের': 256, 'অনুষ্ঠানে': 243, 'দুই': 237, 'উপলক্ষে': 235, 'ছিলেন': 233, 'শুরু': 231, 'হয়': 221, 'উপস্থিত': 218, 'বিতরণ': 218, 'কথা': 207, 'সাথে': 200 |
| Lowest Occurance | 'ঈশ্বরদী': 1, 'সাংবাদিকেরা': 1, 'ইনভেস্টমেন্ট': 1, 'বিনিয়োগকারীদের': 1, 'স্টক': 1, 'সানাউল': 1, 'পরিচালকবৃন্দ': 1, 'শেয়ারহোল্ডার': 1, 'হলি': 1, |

| | 'ফ্যামিলি': 1, 'চালনায়': 1, 'দায়িত্বশীল': 1, 'গণগ্রন্থাগারের': 1, 'জেলাগুলা': 1, 'বাহাদুরপুর': 1, 'পল্লীতে': 1, 'মুট': 1, 'রোভারমুটে': 1, 'দফতরের': 1, 'রিজিয়ন': 1, 'আইনুল': 1, 'বায়েজিদ': 1, 'গণমাধ্যমকর্মী': 1, 'জয়পুরহাটে': 1, 'জালনোট': 1, 'ওয়ার্কশপ': 1, 'যুগ্ম': 1, 'সদস্যগণ': 1, 'পিঠাসহ': 1, |
|---|---|

## 3.7 Vocabulary Restriction

The amount of vocabulary (11359 unique words) compared to a sample size 5642(5130 unique) is huge. Moreover, the sentences are not gloss annotated. The dataset obviously perform poorly in benchmarking. We looked for common vocabulary in train-test-val splits. In a first iteration, we restricted test and val set to retain only those sentence that have vocabulary present in training set. We obtained 509 test samples with vocabulary size 1719 and 254 val set samples with vocabulary 1219. We then restricted the training set to retain only those samples that have vocabulary in common with test-val set. We obtained 781 training samples with vocabulary of 997, meaning (3964-781)= 3,183 samples have a least one new/uncommon vocabulary. We had to iterate over 4 times a reach the common vocabulary split of 810 sample (train-test-val as 567-157-86). We publish the dataset as **IsharaKhobor_SMALL**

Table 4: Statistics of IsharaKhobor_SMALL and IsharaKhobor_CANONICAL_SMALL

| Dataset | #Samples (# unique) | #train-test-val (ratio) | Vocabulary | train-test-val Vocabulary | test-val OOV |
|---|---|---|---|---|---|
| **IsharaKhobor** | 5642 (5130 unique) | 0.70-0.196-0.100 | 11359 | 9636-4230-3148 | 1119-690 |
| **IsharaKhobor_Canonical_SMALL** | 811 (**306 unique**) | 0.69-0.19-0.10 | 308 | 308-268-211 | 0-0 |
| **IsharaKhobor_SMALL** | 810 (312 unique) | 0.7-0.19-0.1 | 313 | 313-269-216 | 0-0 |

## 3.8 Canonicalization

We understood that the variations of vocabulary also contributed to the number of unique vocabulary in the dataset. Spelling variations were mainly introduced by the presenters like মোহাম্মদ (Mohamamd), মুহাম্মদ (Muhammad), আল্লাহ হাফেজ (Allah hafez), খোদা হাফেজ (khoda hafez), হাফিজ (hafiz) etc.

There were still lot of vocabulary in neutral topics as it presents news of the whole country and on a whole bunch of topics. However we found language styles are limited.

We deliberated standardized আসসালামু আলাইকুম (assalamu alaikum) and ইন্না লিল্লাহি ওয়া ইন্না ইলাইহি রাজিউন ( inna lillahe wa inna ilaihe rajiun), irrespective of pronunciation. Still when we executed further standardization 87 of 811 in IsharaKhobor-SMALL were affected.

We canonicalized on a restricted vocabulary and styles as seen in Bangla text pairs below(text, canonicalized) and on the small dataset we obtained, we got some interesting results.

| | | |
|---|---|---|
| "চিত্র নিয়ে", "চিত্রের ভিত্তিতে" | "তথ্য ভিডিও", "তথ্য ও ভিডিও" | "উপর ভিত্তি করে", "ভিত্তিতে" |
| "ভিত্তি করে", "ভিত্তিতে" | "মুহাইমেন", "মোহাইমিন" | "মুহাইমিন", "মোহাইমিন" |
| "খোদা হাফেজ", "খোদা হাফিজ" | "আল্লাহ হাফেজ", "খোদা হাফিজ" | "ঈদ মুবারক", "ঈদ মোবারক" |

The canonicalization operation was done on the whole dataset and common vocabulary splits were obtained in the same manner discussed in sub-section "Vocabulary Restriction**".** We publish this dataset as I**haraKhobor_CANONICAL_SMALL** with 811 samples (567-157-86).

Luckly SMALL and CANONICAL_SMALL have the same test set. Train set varied in two samples and val set varied in one. This allowed the two small datasets comparable. We also observed that if we compare I**haraKhobor**_CANONICAL_SMALL as standard and I**haraKhobor**_SMALL as augmented dataset it is a 10% augmentation in styles and spelling variation.

## 4. Experimental Results

### 4.1 Model and Embeddings

We experimented on two gloss-free transformer architectures namely Sign Language Translation Transformer (SLTT) and Gloss Attention Sign Language Transformer (GASLT).

We used mediapipe landmark based embedding and used only the pose and hand landmarks. These landmarks together obtained 225 points(75x3+2x21x3). To allow experiments with multi-head attention, we made the embedding 224 discarding the last depth point of pose (depth of one of the toe fingers). We call it Raw embedding.

In an effort to have a signer invariant normalized embedding we adapted the embedding to RQE (Relative and Quantized Embedding), which can be found in our previous work [19].

We did an ablation of RQE on a sentence level sign language dataset, PHOENIX-Weather-2014T, and found RQE performs better than GASLT benchmark.

Table 5: Ablation of RQE on GASLT with PHOENIX-Weather-2014T

| *Embedding-Transformer* | *BLEU-1* | *BLEU-2* | *BLEU-3* | *BLEU-4* | *ROUGE* |
|---|---|---|---|---|---|
| *GASLT* | *39.07* | *26.74* | *21.86* | *15.74* | *39.86* |
| *RQE-GASLT* | *39.13* | *26.87* | *20.13* | *16.09* | *40.52* |

4.2 **Results**

We experimented with different hyperparameters : layers, heads, learning rate, patience.

The natural language sentence length is also a hyperparameter, we kept to 67 in big dataset and 30 for small datasets (as big sentences were dropped because of OOV). Table 6 list the summarized configurations for the two transformers.

Table 6: Transformer Configurations

| Max Sequence Length | 1344, video file V007_56457_57800 |
|---|---|
| Max Sentence Length | 67 in sample V097_13696_14387<br>For small datasets length was reduced to 30 |
|  |  |
| SLTT Hyperparameters that we altered | Layer 4<br>Head 8<br>Batch size: 32,<br>Learning rate: 0.0001, Min learning rate $1 \times 10^{-9}$, decrease factor 0.5, patience 15<br>Small dataset allowed layer 6 and patience 80 |
| GASLT Hyperparameters that we altered | Layer 3 (sometime for small dataset)<br>word2vec vocabulary 200k<br>Sentence level cosine similarity matrix learned from IsharaKhobor and then full canonical dataset sharaKhobor_canonical |

Results show that RQE-GASLT performs better in all datasets because of its language similarity to apply on gloss attention to direct the encoder. The predictive gloss duration allows short attention span for a frame. RQE is normalized encoding, can be signer invariant. Table 7 presents the best results in all datasets.

**Table 7: Best results of the datasets**

| Dataset | Best BLEU-4 , ROUGE | Embedding-MODEL |
|---|---|---|
| **IsharaKhobor** | BLEU-4: 3.88, ROUGE: 11.67 | RQE-GASLT |
| IsharaKhobor (CANONICAL) | BLEU-4: 4.32, ROUGE: 11.87 | RQE-GASLT |
| **IsharaKhobor-SMALL** | BLEU-4: 26.81, ROUGE: 50.45 | RQE-GASLT |
| **IsharaKhobor_CANONICAL_SMALL** | BLEU-4: 25.74, ROUGE: 51.26 | RQE-GASLT |

RQE is supposed to perform better in all settings. However, when canonical and RQE were applied together. **IsharaKhobor-SMALL** provided higher BLEU-4: 26.81 than **IsharaKhobor_CANONICAL_SMALL** BLEU-4: **25.74.** We interpreted as the former dataset is a sentence augmented version of the later. So, the result is justified. This suggests **canonicalization** is good, but **with controlled variation** in language should be allowed to better convergence of gradients. In Bigger dataset Canonical performed better BLEU-4: 4.32 with RQE, because, even with canonicalization there were lot of variations in sentences.

In SLTT canonical with Raw embedding performed a bit better which can be explained as Raw embedding as augmentation of input embedding that allows SLTT to converge on canonicalize datasets. It is also because SLTT cannot optimize learning on language decoding and hence it is supposed to optimize encoder embedding that becomes easy with untruncated Raw embedding. However, RQE allowed extra layers (up to 6) and more patience (up to 80) to generate better encoder embedding. This gives us the idea that with better natural language heuristic, RQE can do better in all datasets.

**Table 8: Results of the datasets**

| Dataset (some desc) (best B4, model) | Split (train-test-val) | Result | | | |
|---|---|---|---|---|---|
| | | SLTT | | GASLT | |
| | | Raw | RQE | Raw | RQE |
| **IsharaKhobor** (B4:3.88, RQE-GASLT) | 3964-1108-570 | B1: 7.88, B2: 5.23, B3: 3.79, B4: 2.88 R: 10.82 | B1: 8.88, B2: 6.26, B3: 4.75, B4: 3.74, R:11.75 | B1: 6.89, B2: 4.84, B3: 3.64, B4: 2.92, R: 9.28 | B1: 8.67, B2: 6.35, B3: 4.85, B4: 3.88, R: 11.67 |
| IsharaKhobor (CANONICAL) (B4:4.32, RQE-GASLT) | | B1:9.06, B2:6.38, B3: 4.77, **B4: 3.76** R:11.87 | B1: 8.35, B2: 5.82, B3: 4.31, B4: 3.33 R:11.14 | B1: 6.72, B2: 4.84, B3: 3.76, B4: 3.08 R: 9.18 | B1: 8.98, B2: 6.71, B3: 5.27, **B4: 4.32** R: 11.87 |
| **IsharaKhobor-SMALL** (Subset based on common vocabulary) (B4:26.81, RQE-GASLT) | 567-157-86 | B1: 35.94, B2:30.05, B3: 24.49, B4: 20.05, | B1: 38.54, B2: 32.91, B3: 27.65, B4: 23.76, | B1:33.86, B2:29.33, B3:25.01, B4:21.88 | B1: 40.17, B2: 35.32, B3: 30.63, **B4: 26.81**, |

|  |  | R:49.86 | R: 49.62 | R:42.35 | R: 50.45 |
|---|---|---|---|---|---|
| IsharaKhobor_CANONICAL_SMALL (Canonicalized first, and made subset based on common vocabulary) **(B4: 25.74, RQE-GASLT)** | 567-157-87 | B1:40.33, B2:34.93 B3:29.34 **B4:25.04** R:51.79 | B1: 35.75, B2: 31.43, B3:27.19, B4: 24.16 R:44.73 | B1: 36.91, B2: 32.31, B3: 27.58, B4: 23.93, R:50.83 | B1: 40.63, B2: 35.62, B3: 30.18, **B4: 25.74**, R: 51.26 |

B1: BLEU-1, B2: BLEU-2, B3: BLEU-3, B4: BLEU-4, R: ROUGE

## 5. Discussions

Gloss annotation is known will increase accuracy. But the large scale datasets that can be collected are most from TV news and are without gloss annotations. However, the gloss annotations for sentences can be done independently, without SLT dataset[8]. When the two branches of work in SLT and gloss annotation for Sign Language Production (SLP) i.e. sign language synthesis, will converge, we can expect accurate translations.

So, the way forward can be:

-We do gloss annotations.

- We generate synthetic dataset, especially more synthetic video samples.

- We standardize the variations in spoken language.

Generating synthetic video dataset without gloss annotations are challenging though not impossible. Simultaneously researchers can systematically collect glosses and generate HamNoSys notations which in turn can be used to generate synthetic datasets. One recent work to generate synthetic animation for a gloss can be found in [20].

The third solution is going for standardization of spoken sentences. Actually glosses are mapped to 'limited set of spoken vocabulary styles' and probably SLT cannot generate rich language translations. If a particular translation style and vocabulary are to be generated in SLT they have to be in the training set and have to be dominant over other styles. In this work we experiment with vocabulary standardization i.e. canonicalization in limited scope and report the insight gained.

**Acknowledgement**

This project was generously funded by Information and Communication Technology Division (ICTD) of Ministry of Posts, Telecommunications and Information Technology, People's Republic of Bangladesh.

**References:**

[1] Husne Ara Rubaiyeat, Hasan Mahmud, and Md Kamrul Hasan. (2025). IsharaKhobor [Data set]. Kaggle. https://doi.org/10.34740/KAGGLE/DSV/13878187

[2] Camgoz, N. C., Hadfield, S., Koller, O., Ney, H., & Bowden, R. (2018). Neural Sign Language Translation. *2018 IEEE/CVF Conference on Computer Vision and Pattern Recognition*, 7784–7793. doi:10.1109/CVPR.2018.00812

[3] Zhou, H., Zhou, W., Qi, W., Pu, J., & Li, H. (2021, June). Improving Sign Language Translation with Monolingual Data by Sign Back-Translation. *2021 IEEE/CVF Conference on Computer Vision and Pattern Recognition (CVPR)*, 1316–1325. doi:10.1109/CVPR46437.2021.00137

[4] Duarte, A., Palaskar, S., Ventura, L., Ghadiyaram, D., DeHaan, K., Metze, F., … Giro-i-Nieto, X. (2021). How2Sign: A Large-scale Multimodal Dataset for Continuous American Sign Language. 2021 IEEE/CVF Conference on Computer Vision and Pattern Recognition (CVPR), 2734–2743. doi:10.1109/CVPR46437.2021.00276

[5] Albanie, S., Varol, G., Momeni, L., Bull, H., Afouras, T., Chowdhury, H., … Zisserman, A. (2021). BBC-Oxford British Sign Language Dataset. arXiv [Cs.CV]. Retrieved from http://arxiv.org/abs/2111.03635

[6] Martinez, A. M., Wilbur, R. B., Shay, R., & Kak, A. C. (2002). Purdue RVL-SLLL ASL database for automatic recognition of American Sign Language. *Proceedings. Fourth IEEE International Conference on Multimodal Interfaces*, 167–172. doi:10.1109/ICMI.2002.1166987

[7] Mahbub Zeeon, I. E., Mohammad, M. M., & Adnan, M. A. (2024). BTVSL: A Novel Sentence-Level Annotated Dataset for Bangla Sign Language Translation. *2024 IEEE 18th International Conference on Automatic Face and Gesture Recognition (FG)*, 1–10. doi:10.1109/FG59268.2024.10581873

[8] Saha, N., Shahriyar, R., Roudra, N. A., Sakib, S., & Rasel, A. A. (2025). Introducing A Bangla Sentence - Gloss Pair Dataset for Bangla Sign Language Translation and Research. *arXiv [Cs.CL]*. Retrieved from http://arxiv.org/abs/2511.08507

[9] Bhattacharjee, A., Hasan, T., Ahmad, W., Mubasshir, K. S., Islam, M. S., Iqbal, A., … Shahriyar, R. (2022, July). BanglaBERT: Language Model Pretraining and Benchmarks for Low-Resource Language Understanding Evaluation in Bangla. In M. Carpuat, M.-C. de Marneffe, & I. V. Meza Ruiz (Eds), *Findings of the Association for Computational Linguistics: NAACL 2022* (pp. 1318–1327). doi:10.18653/v1/2022.findings-naacl.98

[10] Sincan, O. M., Low, J. H., Asasi, S., & Bowden, R. (2025). Gloss-free Sign Language Translation: An unbiased evaluation of progress in the field. *Computer Vision and Image Understanding*, *261*, 104498. doi:10.1016/j.cviu.2025.104498

[11] Camgoz, N. C., Koller, O., Hadfield, S., & Bowden, R. (2020). Sign Language Transformers: Joint End-to-end Sign Language Recognition and Translation. *arXiv [Cs.CV]*. Retrieved from http://arxiv.org/abs/2003.13830

[12] Yin, A., Zhong, T., Tang, L., Jin, W., Jin, T., & Zhao, Z. (2023). Gloss attention for gloss-free sign language translation. In *Proceedings of the IEEE/CVF Conference on Computer Vision and Pattern Recognition* (pp. 2551-2562).

[13] Zhou, B., Chen, Z., Clapés, A., Wan, J., Liang, Y., Escalera, S., Lei, Z., & Zhang, D. (2023). Gloss-free Sign Language Translation: Improving from Visual-Language Pretraining. *2023 IEEE/CVF International Conference on Computer Vision (ICCV)*, 20814–20824. doi:10.1109/ICCV51070.2023.01908

[14] Cheng, Y., Wei, F., Bao, J., Chen, D., & Zhang, W. (2023). CiCo: Domain-Aware Sign Language Retrieval via Cross-Lingual Contrastive Learning. *2023 IEEE/CVF Conference on Computer Vision and Pattern Recognition (CVPR)*, 19016–19026. doi:10.1109/CVPR52729.2023.01823

[15] Chen, Y., Zuo, R., Wei, F., Wu, Y., Liu, S., & Mak, B. (2022). Two-stream network for sign language recognition and translation. *Proceedings of the 36th International Conference on Neural Information Processing Systems*. Presented at the New Orleans, LA, USA. Red Hook, NY, USA: Curran Associates Inc.

[16] Kim, J., Jeon, H., Bae, J., & Kim, H. Y. (2025). Leveraging the Power of MLLMs for Gloss-Free Sign Language Translation. *arXiv [Cs.CV]*. Retrieved from http://arxiv.org/abs/2411.16789.

[17] Tariquzzaman, M., Ishmam, M. F., Muna, S. S., Hasan, M. K., & Mahmud, H. (2025). Prompting with Sign Parameters for Low-resource Sign Language Instruction Generation. *arXiv [Cs.HC]*. Retrieved from http://arxiv.org/abs/2508.16076

[18] Rubaiyeat, H.A., Mahmud, H., Habib, A., Hsan, M.K. (2025). BdSLW60: A word-level bangla sign language dataset. *Multimed Tools Appl* **84**, 42399–42423. https://doi.org/10.1007/s11042-025-20792-4

[19] Rubaiyeat, H. A., Youssouf, N., Hasan, M. K., & Mahmud, H. (2025). BdSLW401: Transformer-Based Word-Level Bangla Sign Language Recognition Using Relative Quantization Encoding (RQE). *arXiv [Cs.CV]*. Retrieved from http://arxiv.org/abs/2503.02360

[20] Rahman, M. Shahidur and Islam, MD. Ashikul and Dewan, Prato and Islam, Md Fuadul, Isharakotha: A Comprehensive Avatar-Based Bangla Sign Language Corpus. Available at SSRN: http://dx.doi.org/10.2139/ssrn.4696066